\documentclass[a4paper, 10pt, conference]{ieeeconf}      

\usepackage{balance}
\usepackage{multirow}
\usepackage{graphicx}
\usepackage{hyperref}
\IEEEoverridecommandlockouts                              

\title{\LARGE \bf
Mechanical Chameleons: Evaluating the effects of a social robot's non-verbal behavior on social influence
}

\author{Patrik Jonell$^{1,\dagger}$, Anna Deichler$^{1,\dagger}$, Ilaria Torre$^{1}$, Iolanda Leite$^{1}$, and Jonas Beskow$^{1}$
\thanks{$^\dagger$Both authors contributed equally to this research.}
\thanks{$^{1}$KTH Royal Institute of Technology}
\thanks{*This work was supported by the Swedish Foundation for Strategic Research, project EACare under Grant No.: RIT15-0107 and by the Swedish Research Council project 2018-05409 (StyleBot)}%

}

\begin{document}

\maketitle
\thispagestyle{empty}
\pagestyle{empty}

\begin{abstract}
In this paper we present a pilot study which investigates how non-verbal behavior affects social influence in social robots. We also present a modular system which is capable of controlling the non-verbal behavior based on the interlocutor's facial gestures (head movements and facial expressions) in real time, and a study investigating whether three different strategies for facial gestures (``still'', ``natural movement'', i.e. movements recorded from another conversation, and ``copy'', i.e. mimicking the user with a four second delay) has any affect on social influence and decision making in a ``survival task''. Our preliminary results show there was no significant difference between the three conditions, but this might be due to among other things a low number of study participants (12).

\end{abstract}

\section{INTRODUCTION}

\label{intro}
As humans, we continuously adapt our behavior to the situation we are currently in.  This continuous adaptation is an inherent part of human functioning, and one of the key success factors to human communicative abilities. 
In face-to-face interaction, we reliably adopt poses, facial expressions, mannerisms and speaking styles of the person we are talking to. This phenomenon, known as the chameleon effect, or behavioral mimicry, emerges during infancy and is a fundamental mechanism in how we learn social codes and behaviors, but stays throughout our lives as an important social function \cite{chartrand1999chameleon}. Mimicry has been shown to increase empathy, liking and affiliation between the individuals, and it has been referred to as a social glue \cite{lakin2003chameleon} because it helps to create strong ties and relationships. Mimicry has been shown to increase learning and attention in tutoring situations \cite{martin2012mimicry} as well as to increase trust \cite{maddux2008chameleons}, however, some later studies have had issues with fully corroborating this claim \cite{doi:10.1080/01691864.2019.1589570,hale2017using}. 


In social human-robot interaction, the potential benefits of incorporating behavioral mimicry are clear. For social robot applications in health and elderly care, for example, it is critical that the users trust the robot. Increased engagement, attention and sense of agency on the part of the robot are other desirable effects that could be expected. Much research has gone into generating speech- and text driven non-verbal behavior for robots, but most current human-agent interaction systems do not take the interlocutor into account when generating behaviors, to a large part because such modeling is non-trivial. However, given a reliable method of capturing facial gestures (head movements and facial expressions), copy mimicry with 4 s delay is relatively trivial to implement. Most users do not even notice that they are being copied \cite{bailenson2004transformed}.


In the present contribution, 
we investigate if mimicry of facial gestures has a social influence on participants in a ``survival task'' \cite{Hall1970}  compared to a ``natural movement'' condition (i.e. facial gestures collected from a random conversation) and a ``still'' condition (i.e. a still face). A ``survival task'' is a task where a participant and an agent negotiates necessary items to bring in an emergency situation. The participant will propose a list of prioritized items, while the agent will try to convince the participant to prioritize the list differently. 

To this end we built a system which is capable of real-time interactions with a user through a wizard-of-oz style interface. The system collects facial gestures using an Apple iPhone, similar to \cite{10.3389/fcomp.2021.642633} and \cite{malisz2019visual}, and passes them to a module which in turn generates the social robots facial gestures. In the current study we used a Furhat robot \cite{moubayed2013furhat}.

The main research question of our study is \textit{do different strategies for a social robot to produce facial gestures (in this case the strategies are; ``still'', ``natural movement'', and ``copy'') have an impact on the robot's social influence over the user}.


\section{Related work}
To get an overview of the related work we should review literature in social influence in social agents and methods for measuring this social influence.
\begin{figure*}[t]
\begin{center}
\includegraphics[width=\textwidth]{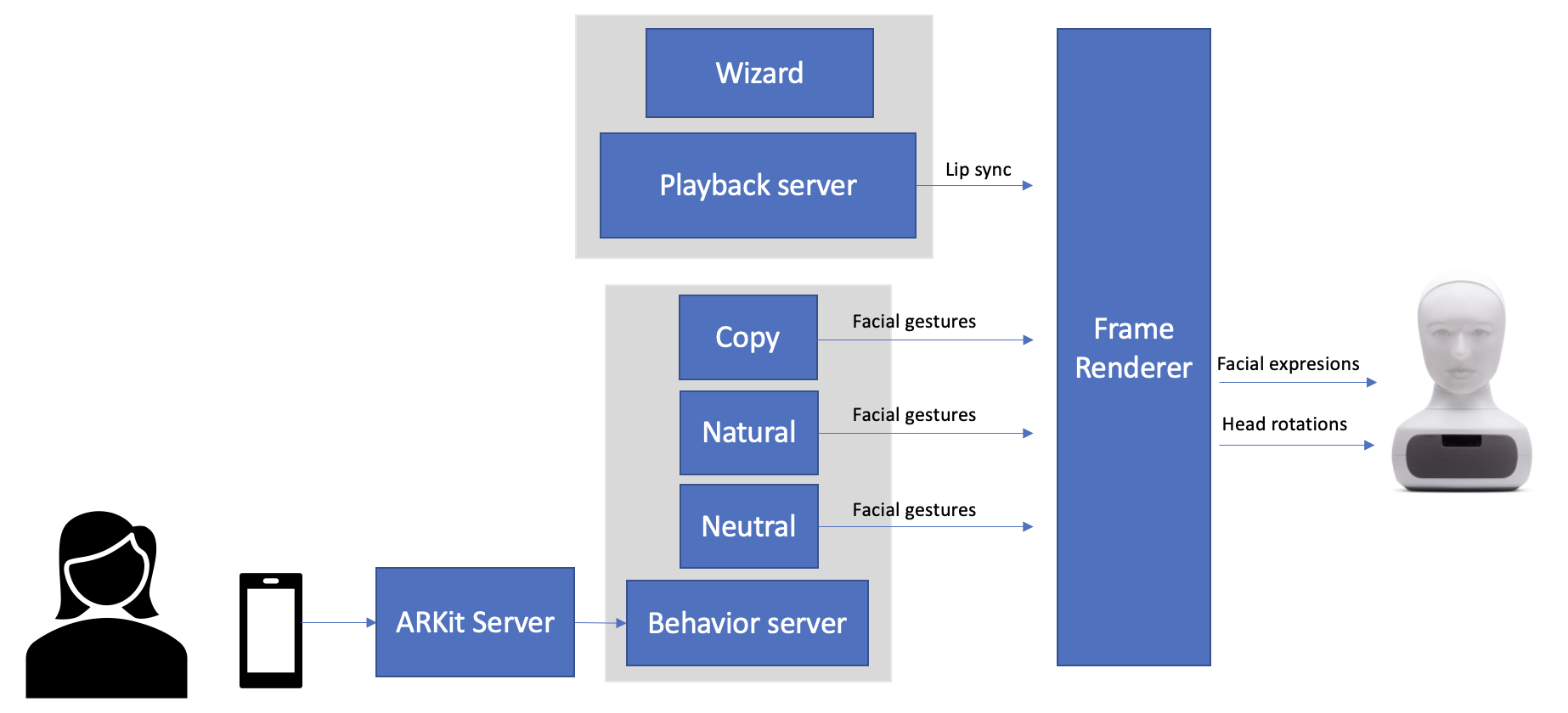}
\caption{Overview of the system architecture showing how the various components interact with each other.}
\label{fig:system_overview}
\end{center}
\end{figure*}
\label{related_work}
\subsection{Social influence in social agents}
Bailenson and Yee investigated how mimicry affects social influence in virtual agents in virtual reality by letting a virtual agent read a paragraph with a persuasive argument, and see if participants agreed or disagreed with the argument. It was found that  having an agent which mimicked the participant's head movements with a 4 s delay was exerting higher social influence i.e. persuaded the participants to a higher degree, than an agent which just natural motion taken from another participant and was also more liked by the participants \cite{doi:10.1111/j.1467-9280.2005.01619.x}. However, several studies has since shown results contrary to those presented by Bailenson and Yee. Ghazali et al. for example did not find such effects when presenting participants with three different persuasive tasks and having a robot mimic the participant's head movements. They did however find that social praise seemed to affect social influence, while head mimicry seemed to affect likeability \cite{doi:10.1080/01691864.2019.1589570}. Hale explored head and torso mimicry in agents presented in virtual reality and did not find it affecting trust, however, similarly to other researchers, the author found mimicry to have a positive effect on the social evaluation of the agent \cite{hale2017using}. None of the works described above consider mimicry of head movements and facial expressions simultaneously, and to the best of our knowledge we are not aware of such works. 


\subsection{Measuring social influence}


There have been various ways social influence has been measured in human-agent interaction settings. For example Ghazali et al. used an experiment with three tasks (one physical task, and two persuasive tasks) involving a participant and a robot. The first of the persuasive task was a picture selection task, where the participants where expected to provide their preference between two pictures, and subsequently describe the picture for the robot. In the second persuasive task the participants were expected to select a card containing a reward \cite{doi:10.1080/01691864.2019.1589570}. These tasks provide only limited amounts of interaction between the agent and the participant.

Another example is Bailenson and Yee who used a ``persuasive passage'', where the agent would present an argument to the participant and ask if they agree or disagree with the argument. However, this does not provide much opportunities for exposing the participants to the agent's behavior strategy \cite{doi:10.1111/j.1467-9280.2005.01619.x}. Some researchers have used investment games, such as Torre et al. \cite{TORRE2020106215}, however investment games like these also tend to have a limited amount of interaction between the participant and the agent.

Hale used a picture description task where the participants and the agent took turns describing pictures to one another. Although this method seems to evoke more conversation between the agent and the participant, there was no persuasive attempt made by the agent \cite{hale2017using}.

Another type of task that has been widely used in measuring social influence in agents is the ``lunar survival task'', also called the ``Lost at the Moon task'' \cite{Hall1970}, used by for example Lucas et al. to investigate normative and informational social influence \cite{cerrors}. In this task participants are presented with a scenario that they are stuck on the moon and asked to rate several objects in terms of their importance for survival. The agent will propose changes to the participant's initial ratings. Using this task it is possible to both have a longer dialog with multiple turns, as the agent and the participant can discuss each object and argue for them, and there is a persuasive element as part of the task. There is also an additional task called the ``desert task'', where the scenario is played out in the desert with different items instead.


.

\section{System architecture}
\label{system_architecture}
The system architecture consists of several modules which can be seen in the schematic diagram in Figure \ref{fig:system_overview}. The first one is the iPhone app, which is a slightly modified version of FaceCaptureX\footnote{\href{https://github.com/elishahung/FaceCaptureX}{https://github.com/elishahung/FaceCaptureX}}, which streams blendshapes and head rotations in real-time via a sockets connection. The app was modified to also send timestamps with each frame, additionally some performance improvements were implemented. The data from the Apple iPhone is received in a module called the ARKit Server which parses the data into JSON format, renames the blendshapes to conform with the furhat's blendshape names, and rotates the z axis 90 degress CCW. The ARKit Server then publishes the data using ZeroMQ\footnote{\href{https://zeromq.org/}{https://zeromq.org}} to be received by the behavior to server. The behavior server dispatches the data to the correct submodule, depending on what behavior is currently set. These submodules take the facial gestures from the user as an input and publish agent behavior over ZeroMQ as a result. The system is setup so that it is easy to add more submodules with various agent behaviors. 
The playback server receives commands from the wizard regarding which sound to play and plays the sound of one of the pre-recorded prompts, and publishes the blendshapes relevant for the lipsync over ZeroMQ. These pre-recoded prompts were recorded using Live Link Face\footnote{\href{https://apps.apple.com/us/app/live-link-face/id1495370836}{https://apps.apple.com/us/app/live-link-face/id1495370836}}.
The Frame Renderer has two loops which are running constantly. One which receives blendshapes over ZeroMQ, and saves that into a state variable, and the other which runs at a constant frame rate (125fps) and reads from the state variable. This module also applies some smoothing to the head rotations using a finite impulse response (FIR) filter and sends the commands for either changing blendshapes or rotating the servos at 25 and 125 fps respectively.

.

\section{Experimental setup}
\label{experimental_setup}
We carried out an experiment to investigate the effect different robot behavior strategies have on social influence in a human-robot interaction setting. The experiment was based on the lunar survival task \cite{Hall1970}. There are various implementations of the lunar survival task, such as \cite{Werkhoven2001SeeingIB,10.1145/1979742.1979604,Artstein2017}, but for this experiment the agent would describe each object, and prompt the participant about what their opinion about those objects were. It would then ask them to rank the objects on a graphical interface on a phone. Finally, the agent would request to change the order of three items such that it would better conform to a pre-determined optimal order, established by NASA \cite{Hall1970}. If the order already was at its optimal configuration or only objects that the participant has declined are in an suboptimal position, the agent would randomly choose an item to move to a position above its current position.
The dependent variable of this experiment is how many times the participants accepted the robot's proposed changes.

Each participant interacted with three robot personalities, and each of the personalities had different strategies when it came to facial gestures (i.e. head movements and facial expressions). The three strategies were: ``still''; where the robot did not move it's head nor did any facial expressions, ``natural movement''; which was a recording of facial gestures from an unrelated conversation, and finally the ``copy'' condition, where the robot is copying the facial gestures with a 4s delay, as proposed in \cite{bailenson2004transformed}. The experiment was a wizard-of-oz style experiment, and had two scenarios, a lunar scenario and a desert scenario, both of which had lists of 15 objects with experimentally assessed relative importance between the objects defined. Since there were three conditions, we selected ten items from the middle of the list, and from these ten items we selected a set of ``better'' items (the upper five), and a set of ``worse'' items (the lower five), yielding four scenarios. 
\begin{figure}[t]
\begin{center}
\includegraphics[width=\columnwidth,height=300pt]{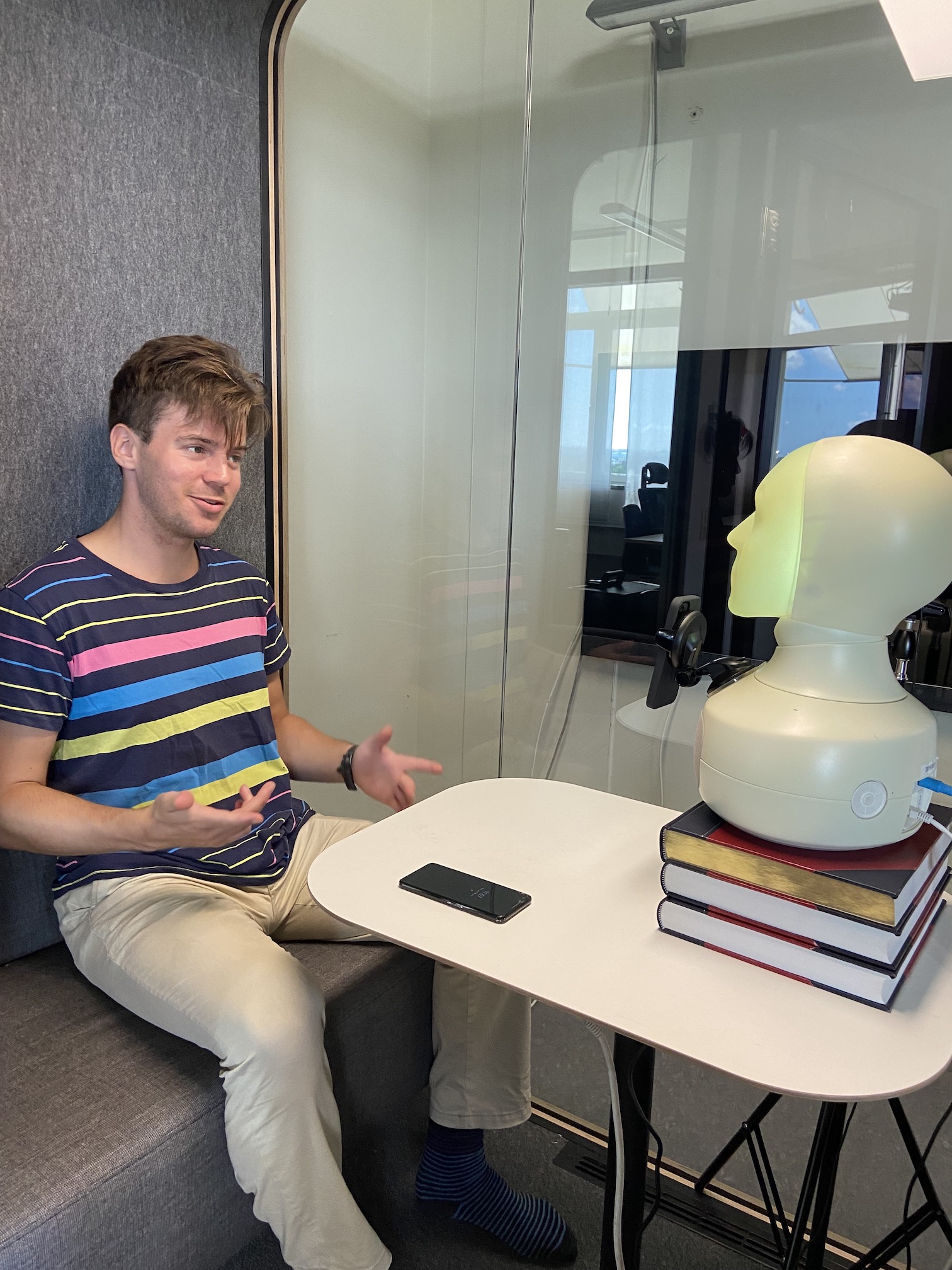}
\caption{The participants were interacting with the Furhat robot in a soundproof booth. On the table there was also a phone for rating the objects during the interaction.}
\label{fig:interaction}
\end{center}
\end{figure}
For this experiment we created four robot personalities, as we wanted each interaction to be with a distinctly different robot. We used a different person for each personality as an actor (two male, and two female), and recorded 232 prompts per actor. Static voice prompts were used since one of our behavioral submodules (outside of the scope for this manuscript) required that the agent had a real human voice as opposed to a text-to-speech (TTS) voice. Each actor was given a randomized name, and we also created four different faces for the Furhat.

As the number of participants was low, We used a within-subjects design for the experiment, such that each participant would interact with each condition of facial gesturing. Counterbalancing was performed as follows: each condition was presented with each actor an equal amount of times, and it was made sure that each condition appeared first, middle, and last an equal number of times. For each experiment only three pairs of condition-actor were chosen, so it was made sure that each actor would have been used an equal amount of times throughout all experiments. This also made sure that there was a balance between robot personalities that were male and female. It was made sure that each of the scenarios would be shown an equal number of times, and that the same task type (i.e. lunar or desert) were never presented twice in a row. It was made sure that the better and worse object sets were presented with both task types an equal number of times across all experiments. Additionally, the robot faces were counterbalanced so that they would appear an equal amount of times.
The objects in each survival task were randomly ordered, however it was made sure that each object at least appeared first and last throughout all experiments.

The experimented was performed in a soundproof booth with the Furhat robot placed on a table in front of the participant (please see Figure~\ref{fig:interaction}). There was one iPhone attached to the Furhat which streamed facial gestures from the participant to the system (see Section~\ref{system_architecture}) and one phone placed on the table used for the participant to enter their initial ranking of the objects. Furthermore there was a webcam on a tripod on the side of the robot facing the participant, which was used for hearing the conversation in the room, and recording audio and video of the participant from the interaction. Next to the soundproof booth, in a separate room, was a work station used as the wizard-of-oz control station.

\subsection{Experiment procedure}
The participants were greeted and asked to fill out a consent form and read an introduction text. This text informed them about the experiment procedure and that they would be asked to interact with the robot three times and then after each interaction rate it using a given questionnaire (same as in \cite{RAU2009587}, except one question added in the end ``Please describe your experience with the robot!''). The questionnaire contained questions relating to three topics; credibility, likeability, and trust. They were asked to familiarize themselves with the questionnaire before the experiment. They were also instructed not to interact with the agent during the rating phase. The reason behind this was that the copy mimicry would be perceived as having joint attention on the phone while the user was rating the objects, while the other conditions would be unable of such behavior. At the end of the experiment the participants were asked to fill out a final questionnaire with one question asking what differences they perceived between the three robots and a second question asking them for any further comments. The experiment took roughly 45 minutes and the participants were awarded a gift certificate valid at a variety of stores worth approximately 15€. 
We recruited 12 participants (4 female, 8 male).

\section{Results}
\label{results}
\subsection{Statistical analysis}
The results were analyzed using a Friedman test, and the p-values were corrected using Bonferroni correction for all tests presented below. The sum of accepted changes (0-3) were compared among the three conditions  ``still``, ``natural'', and ``copy'' ($\chi^2=0.25$, $p=1$), however the results were not statistically significantly different. No post-hoc test was therefore performed. Please refer to Table~\ref{tab:stats} for median and average values for each condition.

The questionnaire (same as in \cite{RAU2009587}) had multiple questions on three topics, credibility (12 questions), likeability (10 questions), and trust (6 questions). The answers from each topic were averaged into three scores as described in \cite{RAU2009587}. No statistical difference were found between the conditions among any of the three topics; credibility ($\chi^2=1.56$, $p=1$), likeability ($\chi^2=1.96$, $p=1$), and trust ($\chi^2=0.79$, $p=1$). No post-hoc test was performed since the differences were not statistically different.

\begin{table}[]
    \centering
    
    \caption{Table showing the median and average over how many times the participants were willing to accept the robots proposition (0-3) and the median and average score for the three topics from the questionnaire, rated on likert scales (1-7).}
    \begin{tabular}{ |c|c|c|c| } 
    \hline
    Condition & Category & Median & Average \\
    \hline
    \multirow{3}*{Accepted robot suggestions} 
    & Natural movement & 1.0 & $1.4\pm0.9$ \\ 
    & Copy & 1.0 & $1.5\pm0.9$ \\ 
    & Still & 1.5 & $1.6\pm0.9$\\ 
    \hline
    \multirow{3}*{Credibility}
    & Natural movement & 4.8 & $4.7\pm1.0$ \\
    & Copy & 5.1 & $4.8\pm1.0$\\ 
    & Still & 4.8 & $4.7\pm1.1$ \\ 
    \hline
    \multirow{3}*{Likeability} 
    & Natural movement & 4.3 & $3.9\pm1.2$ \\ 
    & Copy & 4.2 & $4.2\pm0.9$ \\ 
    & Still & 4.0 & $3.9\pm1.3$ \\ 
    \hline
    \multirow{3}*{Trust}
    & Natural movement & 4.9 & $4.9\pm0.7$\\ 
    & Copy & 5.0 & $4.9\pm1.0$ \\ 
    & Still & 5.1 & $5.0\pm1.5$\\ 
    \hline
    \end{tabular}
    \label{tab:stats}
\end{table}


    
    
\subsection{Participant comments}
\label{subjective evaluations}
Below we present the various comments attributed to each behavior strategy that the participants wrote at the end of each questionnaire.
\subsubsection{Comments on ``copy'' condition} ``seems a bit unfriendly, but felt most intelligent'', ``[the other robots] seemed kinder'',  ``was better with head movement and seemingly reacting to my head movements. It gave me some answers'', ``more attractive to me as it is more warm and friendly.  Overall, I mostly like [the ``copy''] robot'', ``seemed more professional'', ``is full of body/head motion, but it is not open for others'', ``was most life-like and sincere'', ``moved the head actively, was looking to the sides - quite realistic which made them approachable.'' , ``[This] robot left a unsettling feeling. Felt like I was being tricked''

\subsubsection{comments on ``still'' condition} perceived "less intelligent and sympathetic", "appeared much more intelligent both in terms of speech and gestures.",  "not open for others", "[this] one looks like a more strict person, as it does not have many emotions and physical movements', "I did not feel listened to and the robot felt more like a video. The voice was also very stale", "was more serious, similar to what one would expect from an expert.", "was kinder", "robot was never looking directly at me, which made the interaction seem a bit strange"

\subsubsection{comments on ``natural movement'' condition} "[this] one had rather rigid facial expressions, and I was rather disappointed and bored by the interaction.", "[this] robot was more inviting and friendly than the others", "[this] one is the one most open and welcome among these 3, I like it but it feels like not so intelligent", "nicer way of interacting, it was nicer to talk to"

\section{Discussion}
In its current form, our results seems to concur with the results of the studies not being able to establish a link between social influence and head movement mimicry \cite{doi:10.1080/01691864.2019.1589570,hale2017using}. However, the present study is to be seen as a pilot study as the sample size was very low (12). When analyzing the comments that the participants left, it is difficult to discern specific attributes for the three conditions, however it seems like the copy condition was perceived to be more life-like and animated, and also associated with more positive comments than the others. The still condition seemed to be associated with strictness and professionalism, which might explain why participants to a slightly higher degree accepted suggestions from the still condition, as opposed to the other two (however, the statistical tests do not provide evidence for this), and also rated it slightly higher in terms of trust. Furthermore we can see that participants rated the copy condition as slightly more credible and the natural movement condition as slightly more likeable.

One issue that might have affected the results of the study was that due to the complexity of the dialog and the wizard-of-oz setup. There are several aspects of the wizarded interaction which might have interfered and affected the robot's effects on the social influence on the participants. Several participants commented on the robot's behavior not being appropriate to the given emergency situation in the survival tasks, as one of the robot personalities were perceived as too positive. This relates to contextual incongruence, which  have been shown to have  profoundly harmful effects on likability and believability \cite{contxing}.  This is difficult to account for in the copy and natural movement conditions, but should be considered in future studies. Previous studies have also shown that conversational errors (e.g. asking users to repeat themselves, answering different questions, repeating the same answer, not answering questions) can also have a detrimental effect on social influence, making the robot less capable of influencing people as opposed to a robot that do not make errors \cite{cerrors}. As the participants could ask the robot whatever they wanted, and the prompts were pre-recorded, this led to situations were the robot could not answer questions which the participants expected the robot to be able to answer (e.g. how far is 200 miles in kilometers, what are we already equipped with?, etc.). Another issue with having pre-recorded prompts was that they had a certain prosody. We did record variations of some of the most common prompts, and a random variation was chosen each time, but this sometimes led to the participant perceiving the robot as being sarcastic, just from the prosody of the uttered prompt in the given context.

Finally there were sometimes issues with the ranking, as the system tried to optimize the ranking to be as correct as possible according to the pre-determined correct order. If optimal order was already achieved then the system would chose one object at random and ask the participant if they wanted to reorder that item. In some cases when this happened the participants thought it was strange that the robot would suggest a worse item to be put higher on the list.

\subsection{Future work}
Using copy-mimicry as presented in this paper is a simple way of creating dynamic behaviors that are dependent on the interlocutor. However, human-human interaction is more than just mimicking your conversational partner, and a more advanced model could perhaps provide a richer and more interesting interaction. For such interactions a probabilistic generative model, as presented by Jonell et al. \cite{jonell2020let} could for example be used, and compared to the models evaluated in this paper. Since the proposed system architecture allows for adding submodules with robot behavior, a future work direction could be to investigate how such a generative model would compare against for example the copy mimicry strategy.


\section{Conclusion}
In this paper we have presented a pilot study which tries to investigate whether different strategies for generating facial gestures has an impact on social influence. The three strategies evaluated was a still head, with no facial expressions more than the lip sync, a ``natural movement'' condition where facial gestures have been recorded during a random interaction, and a delay copy mimicry condition where the participants facial gestures were copied with a 4 s delay. From the comments of the participants it is clear that the mimicry behavior makes several users perceive the robot as more competent and less rigid, although the ratings-based metrics did not show a clear difference. We believe that more research is needed to reach a conclusions on the effect of the different behavior paradigms. 

\addtolength{\textheight}{-12cm}   





\section*{ACKNOWLEDGMENT}
The authors would like to thank Alireza Mahmoudi Kamelabad, Birger Mo\"ell, and Ulme Wennberg for helping out with testing the experiment.

\balance
\bibliography{refs}
\bibliographystyle{plain}

\end{document}